\documentclass[11pt,a4paper]{article}
\usepackage[hyperref]{emnlp-ijcnlp-2019}
\usepackage{times}
\usepackage{latexsym}
\usepackage{array}

\usepackage{url}

\usepackage{wrapfig}
\usepackage{amsmath}
\usepackage{graphicx}
\usepackage[colorinlistoftodos]{todonotes}
\usepackage{tikz}
\usetikzlibrary{decorations.pathreplacing}
\usetikzlibrary{shapes.multipart}
\usetikzlibrary{arrows.meta}

\usepackage{algpseudocode}
\usepackage[ruled]{algorithm2e}

\definecolor{niceblue}{HTML}{0074D9}
\usepackage{marginnote}
\DeclareMathOperator*{\argmax}{arg\,max}
\DeclareMathOperator*{\argmin}{arg\,min}

\usepackage{subcaption}
\usepackage{booktabs}
\usepackage{makecell}
\usepackage{multirow}
\usepackage{footnote}
\usepackage{bm}
\makesavenoteenv{tabular}
\makesavenoteenv{table}

\usepackage{pifont}
\newcommand{\cmark}{{\large\ding{51}}}%
\newcommand{\xmark}{{\large\ding{55}}}%

\aclfinalcopy 

\title{The \emph{Shmoop} Corpus:\\A Dataset of Stories with Loosely Aligned Summaries}

\author{Atef Chaudhury$^{1,2}$, Makarand Tapaswi$^{1,2,4}$, Seung Wook Kim$^{1,2,3}$, Sanja Fidler$^{1,2,3}$\\
$^1$University of Toronto, $^2$Vector Institute, $^3$NVIDIA, $^4$Inria \\
\url{http://www.cs.toronto.edu/~makarand/shmoop/}\\
{\small \texttt{\{atef,makarand,seung,fidler\}@cs.toronto.edu}} }

\date{}

\begin{document}
\maketitle
\begin{abstract}
Understanding stories is a challenging reading comprehension problem for machines as it requires reading a large volume of text and following long-range dependencies.
In this paper, we introduce the \emph{Shmoop} Corpus: a dataset of 231 stories that are paired with detailed multi-paragraph summaries for each individual chapter (7,234 chapters), where the summary is chronologically aligned with respect to the story chapter.
From the corpus, we construct a set of common NLP tasks, including Cloze-form question answering and a simplified form of abstractive summarization, as benchmarks for reading comprehension on stories.
We then show that the chronological alignment provides a strong supervisory signal that learning-based methods can exploit leading to significant improvements on these tasks.
We believe that the unique structure of this corpus provides an important foothold towards making machine story comprehension more approachable.

\end{abstract}

\section{Introduction}

Humans have an extraordinary capability to become immersed in long narrative texts such as novels and plays, relive the stories, sympathize with characters, and understand the key messages that authors intend to convey.
This set of skills defines the core of reading comprehension, a problem that remains unsolved by machines.

What makes story comprehension challenging is the fact that novels typically feature very long texts written in a variety of styles, and contain long-range dependencies over characters and plot events.
Existing datasets such as NarrativeQA~\cite{kovcisky2018narrativeqa} consist of stories with a set of questions and answers as a proxy task for understanding.
However, such quizzes provide a very weak form of supervision for an artificial agent, as it is required to learn to extract relevant semantic information from an extremely large volume of text with only a few QA pairs.
This is likely one of the reasons that existing methods on summarization~\cite{see2017get, paulus2017deep,WenyuanArxiv16} and question answering~\cite{hermann2015teaching, hill2015goldilocks, rajpurkar2016squad} are often limited to texts with only a few sentences or paragraphs.
Recent forays into analyzing longer texts~\cite{chen2017reading, joshi2017triviaqa, liu2018generating} focus on Wikipedia articles or search results which are typically simpler in structure and are fact-based, and thus do not require a high-level of semantic abstraction in order to comprehend.

\begin{figure}[t]
\centering
\includegraphics[width=1.0\linewidth]{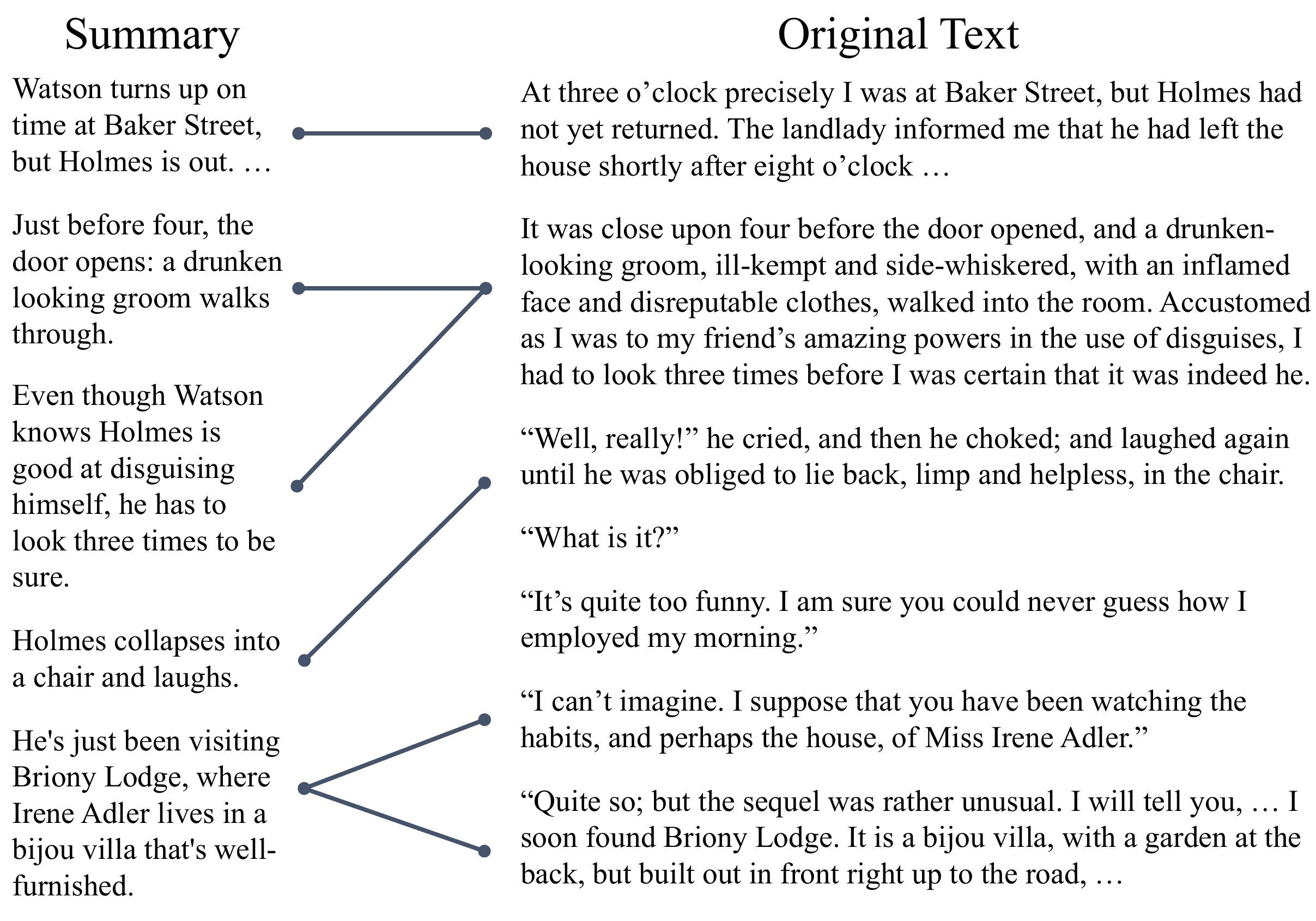}
\vspace{-7mm}
\caption{\label{firstpage-figure} An excerpt from the corpus showing Shmoop summary paragraphs (left) and their chronological alignment to story paragraphs (right).}
\vspace{-2mm}
\end{figure}

To address this, we introduce the \textit{Shmoop} Corpus: a dataset of stories (books and plays) with summaries from Shmoop\footnote{https://www.shmoop.com/}, a learning guide resource.
Rather than summarizing the entire story in a few paragraphs as in Wikipedia plot synopses, these summaries are written for each chapter (see Fig.~\ref{firstpage-figure}).
Each summary compresses the relevant events in the chapter into a few short paragraphs, and is written in a neutral style.
Moreover, the paragraphs in the summaries have a loose chronological alignment with paragraphs in the chapter.
Just as with human learners for whom these summaries are originally intended, we believe that the alignment also provides a strong supervisory signal for training machine comprehension models.

We showcase this by constructing a benchmark set of NLP tasks from the corpus: Cloze-form question answering and a simplified form of abstractive summarization with multiple choices.
We first demonstrate how the chronological structure can help compute alignments.
Then, we use this alignment to learn semantic representations of stories that perform well on the tasks, thus demonstrating that this corpus is a key step towards improving reading comprehension for stories.

\begin{table}[t]
\centering
\small
\begin{tabular}{lccc}
\toprule
\multirow{2}{*}{\bf Corpus} & \multirow{2}{*}{\bf Size} & \multicolumn{2}{c}{\bf Avg Tokens} \\
 & & Summary & Text \\
\midrule
\makecell[l]{CNN/Daily Mail \\ \cite{hermann2015teaching}} & 300k & 56 & 781 \\
\makecell[l]{Children's Book Test\footnote{Corpus does not contain summaries. Size is based on the number of contexts which are derived from 108 texts.} \\ \cite{hill2015goldilocks}} & 700k & -NA- & 465 \\
\makecell[l]{NarrativeQA \\ \cite{kovcisky2018narrativeqa}} & 1,572 & 659 & 62,528 \\
\makecell[l]{MovieQA\footnote{Dataset based on movie scripts and plot summaries.} \\ \cite{MovieQA}} & 199 & 714 & 23,877 \\
\midrule
Shmoop Corpus (Ours) & 7,234 & 460 & 3,579 \\
\bottomrule
\end{tabular}
\vspace{-2mm}
\caption{\label{c_stats} Statistics for summary and narrative datasets.}
\vspace{-2mm}
\end{table}

\section{Corpus}
We describe the features of the \emph{Shmoop} corpus.

\vspace{1mm}
\noindent\textbf{Data collection.}
We first retrieved paired summaries and narrative texts from the Shmoop website and Project Gutenberg\footnote{Accessed at \url{https://www.gutenberg.org/}}, respectively.
The stories were parsed manually, split into chapters (plays into scenes), and then matched to their summaries.
To assess the chronological order between the summaries and stories, we manually labeled a small validation set (5\% of the corpus) with alignments.
To do this, we split summaries into paragraphs based on their bullet-point structure, and stories based on line breaks (up to 250 words).
We then manually aligned the summary and story paragraphs to best match their content.

\vspace{1mm}
\noindent\textbf{Statistics.}
The corpus consists of 231 works (145 novels, 62 plays and 24 short stories) with a total of 7,234 chapter and summary pairs.
Table~\ref{c_stats} compares our dataset to other narrative or summarization datasets.
Our \emph{Shmoop} corpus strikes a balance between short-form large-scale datasets such as the Children's Book Test and long-form small-scale corpora like NarrativeQA.
At the paragraph level, our dataset has 111k summary paragraphs with 30 words on average, and 436k story paragraphs with an average of 59 words each.

\vspace{1mm}
\noindent\textbf{Chronological structure.}
The manually aligned validation set contains 360 chapters (from 15 stories) with a total of 13.5k aligned summary paragraphs.
During alignment, we did not impose any constraints and allowed annotators to skip story paragraphs that did not fit any summary paragraph.
Despite this, only 168 (1.2\%) of the aligned summary paragraphs deviate from chronological order.

\begin{figure*}[t]
\centering
\includegraphics[width=0.192\linewidth]{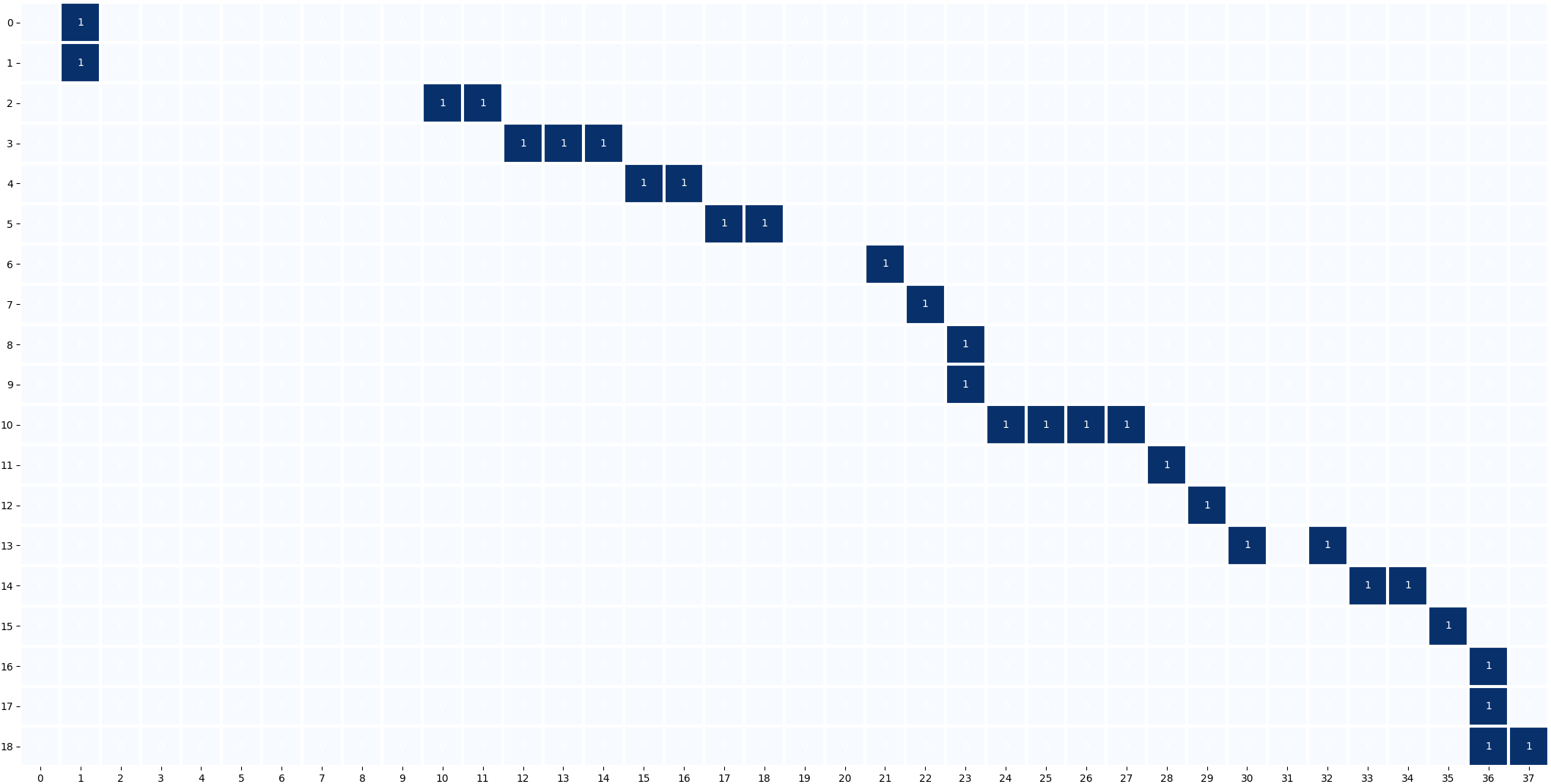}\hspace{0.5mm}
\includegraphics[width=0.192\linewidth]{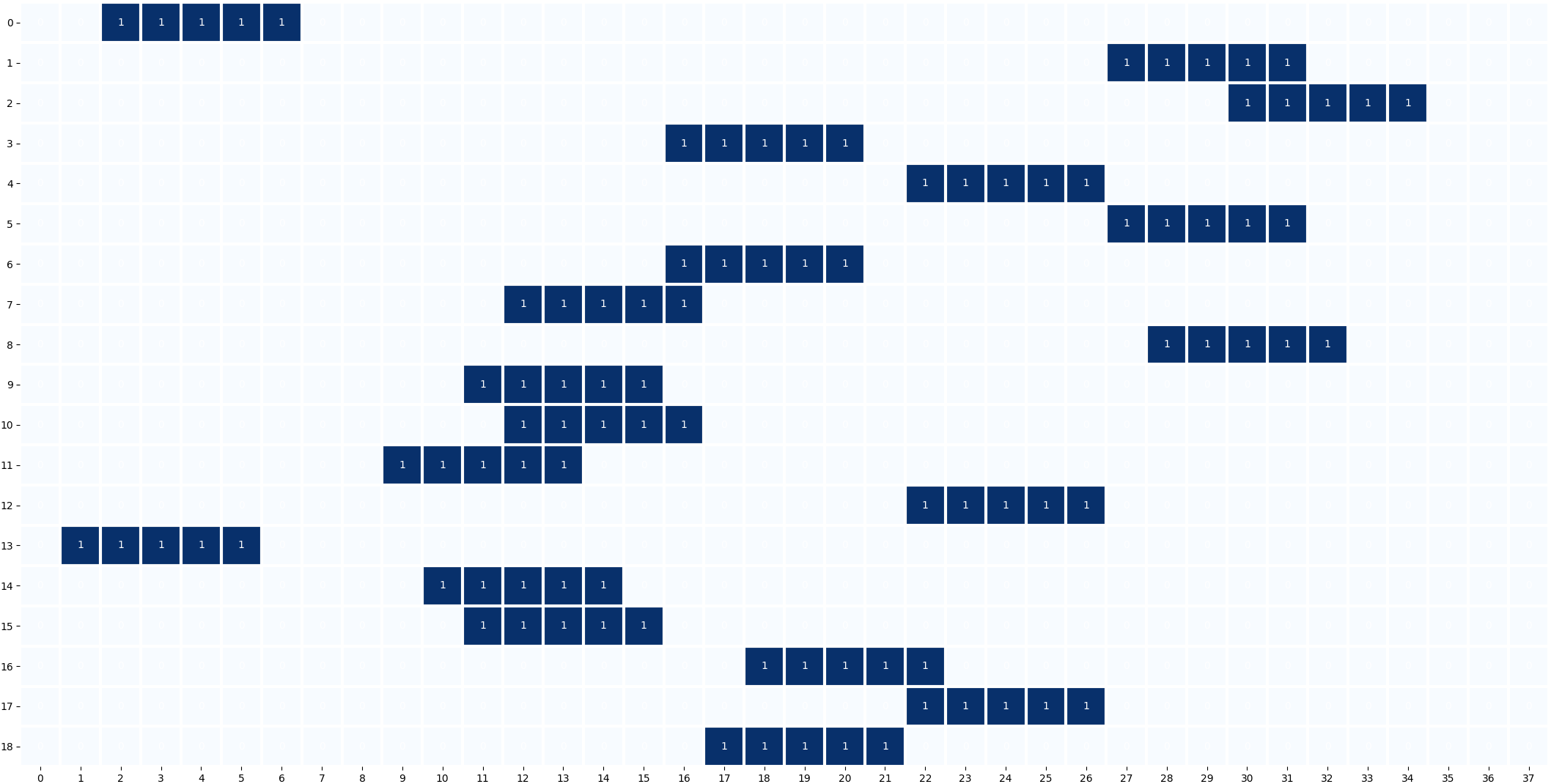}\hspace{0.5mm}
\includegraphics[width=0.192\linewidth]{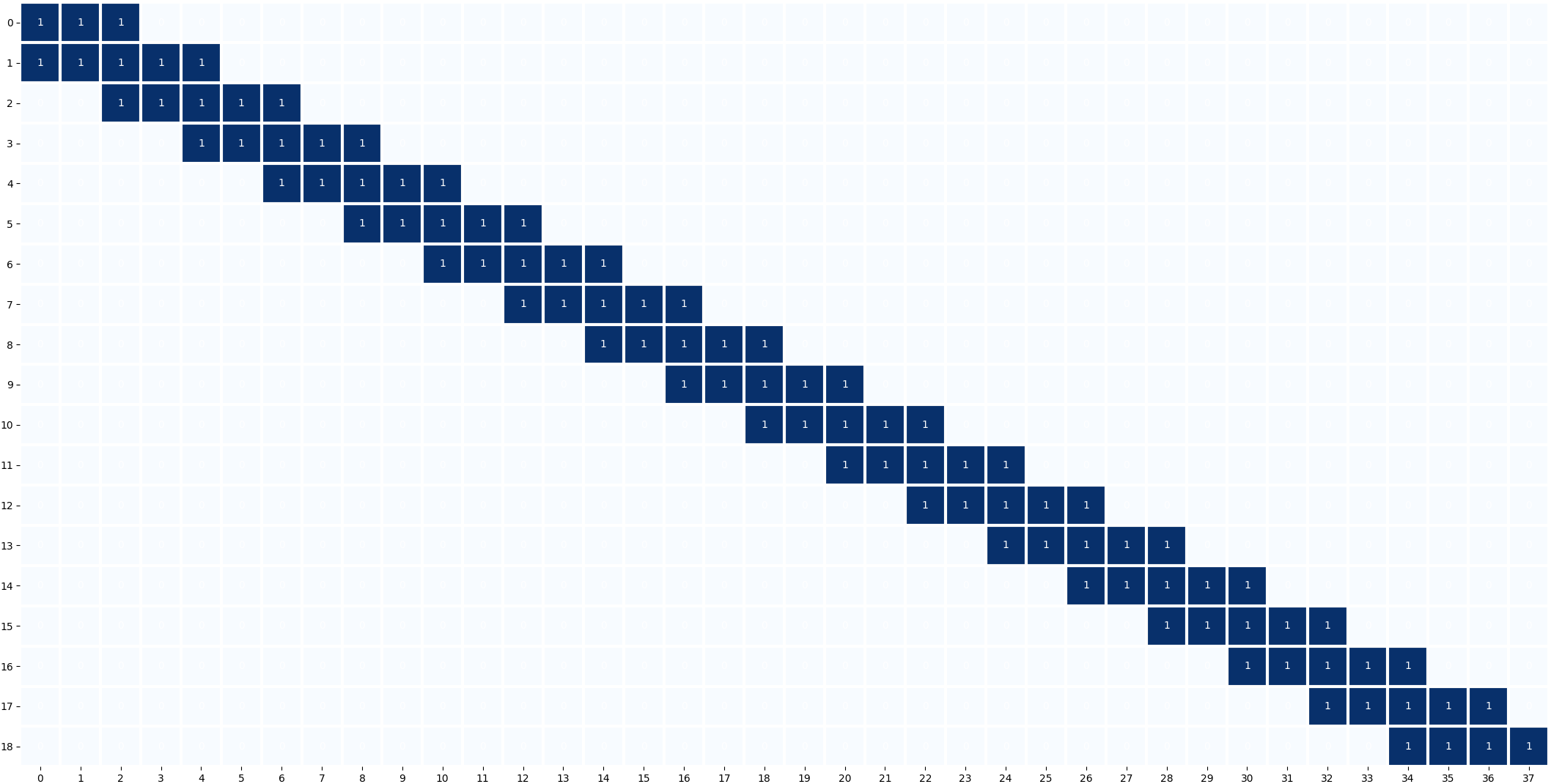}\hspace{0.5mm}
\includegraphics[width=0.192\linewidth]{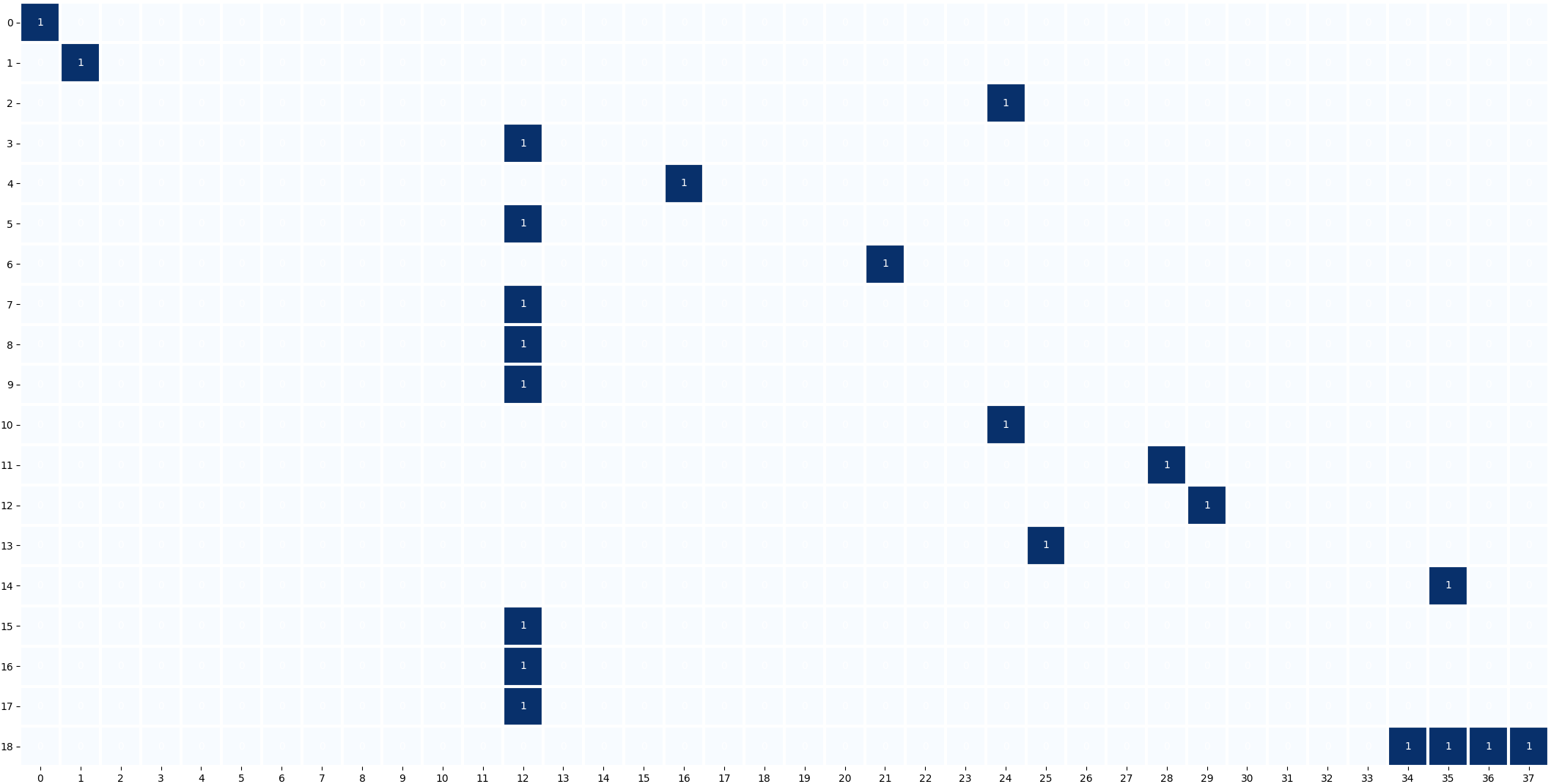}\hspace{0.5mm}
\includegraphics[width=0.192\linewidth]{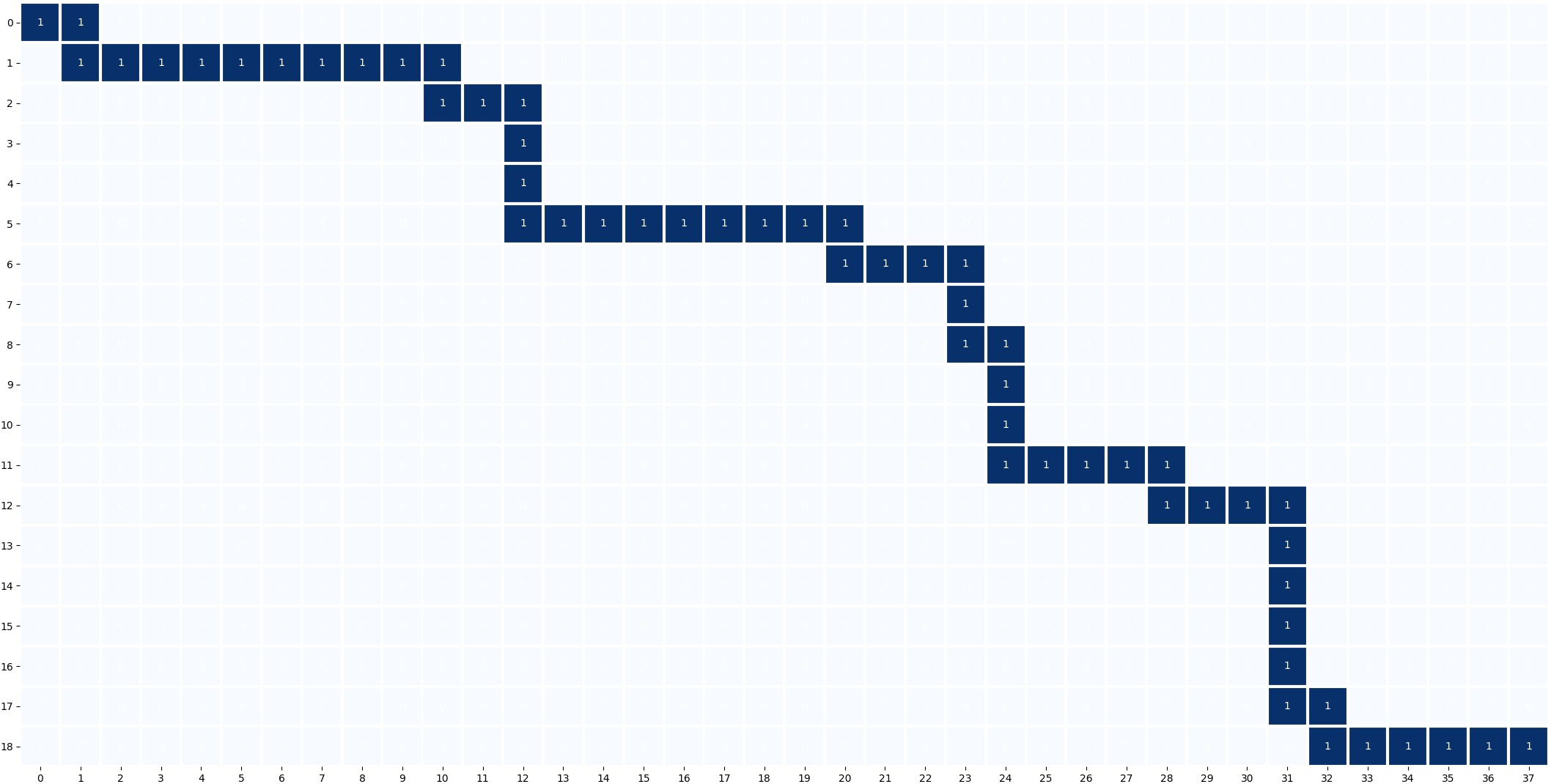}
\vspace{-3mm}
\caption{
Various alignment methods applied to a chapter from \textit{A Midsummer Night's Dream}.
In each plot, y-axis (top to bottom) indicates subsequent summary paragraphs and x-axis (left to right) represents story paragraphs.
A darkened square indicates that the summary paragraph is aligned to the story paragraph.
\textbf{Alignment methods (left-to-right):} Ground-truth, Random-5, Diagonal-5, TFIDF, Chronological-TFIDF.}
\label{fig:alignments}
\vspace{-2mm}
\end{figure*}

\section{Exploiting the Chronological Structure}
We take advantage of the chronological structure in a two step process.
First, we use it to compute alignments that demonstrate good performance on our validation set.
Second, we use these alignments as supervision for our benchmark tasks and show improved performance.
Let us define some notation.
For any chapter, let $S = \{s_i\}_{i=1}^N$ denote the summary composed of a sequence of paragraphs $s_i$, and $D = \{d_j\}_{j=1}^M$ denote the paragraphs of the story document.
An alignment $H$ denotes a relationship between $S$ and $D$, where $h_{ij} \in \{0,1\}$, such that $h_{ij} = 1$ iff $d_{j}$ is \emph{aligned} to $s_{i}$.
Note that by \emph{aligned}, we mean that the summary paragraph $s_i$ partially or wholly encapsulates the content conveyed in the story paragraph $d_j$.

\subsection{Computing Alignments}


Let $g(s_{i}, d_{j})$ denote a scoring function indicating similarity between $s_{i}$ and $d_{j}$.
We define the optimal alignment for a given $g$ as:
$H^{*} = \argmax_H \sum_{ij} h_{ij}\cdot g(s_{i}, d_{j})$,
subject to a set of constraints.
Specifically, we impose chronological ordering with the following constraints: $\forall i>k$ and $j<l$, if $h_{kl} = 1$ then $h_{ij} = 0$.
Similar to prior work on aligning subtitles with scripts~\cite{everingham2006} or plot synopses sentences with video shots~\cite{tapaswi2015ijmir} we can compute alignments efficiently with dynamic programming.

We analyze a variety of alignments with and without imposing the temporal ordering.
See Fig.~\ref{fig:alignments} for an illustration of each approach.

\vspace{1mm}
\noindent\textbf{Random-$\bm{N}$.}
Not considering the similarity score $g$ leads to a random alignment.
Here, $N$ randomly selected consecutive $d_j$ are aligned to each $s_i$.

\vspace{1mm}
\noindent\textbf{Diagonal-$\bm{N}$.}
Imposing temporal order without considering similarity scores leads to a diagonal alignment.
In particular, $N$ story paragraphs are aligned with each summary paragraph.

\vspace{1mm}
\noindent\textbf{Similarity scores.}
We assign $d_{j^*}$ to a summary paragraph $s_i$ such that $j^* = \argmax_j g(s_i, d_j)$.
We compare three different scoring functions $g$.
(i) BLEU~\cite{papineni2002bleu}, a classic metric used in translation;
(ii) BERT~\cite{devlin2018bert}, we compute cosine similarity between paragraph representations extracted with the pre-trained BERT model; and
(iii) TFIDF, compares paragraphs using words weighted by their frequencies.

\vspace{1mm}
\noindent\textbf{Chronological similarity scores.}
Alignment $H^{*}$ is computed using dynamic programming with different scoring functions from above.

\begin{table*}[t]
\tabcolsep=0.12cm
\centering
\small
\begin{tabular}{l cc ccc cc}
\toprule
\multirow{2}{*}{\bf Alignment Type} & \bf Using & \bf Temporal & \multicolumn{3}{c}{\bf Alignment Performance} & \multicolumn{2}{c}{\bf Task Accuracy} \\
& \bf Similarity $g$ & \bf Order & \bf Precision & \bf Recall & \bf F-Score & \bf Cloze & \bf Summarization \\
\midrule
No Context (Random Choice) & N/A & N/A & N/A & N/A & N/A & 0.109 & 0.102 \\
No Alignment (All Paragraphs) & \xmark & \xmark & 0.127 & \textbf{1.000} & 0.206 & 0.149 & 0.148  \\
\midrule
Random-1  & \xmark & \xmark & 0.125 & 0.048 & 0.065 & 0.173 & 0.118  \\
Random-5  & \xmark & \xmark & 0.125 & 0.217 & 0.145 & 0.255 & 0.129  \\
Random-25 & \xmark & \xmark & 0.121 & 0.606 & 0.186 & 0.127 & 0.144  \\
\midrule
Diagonal-1  & \xmark & \cmark & 0.210 & 0.087 & 0.115 & 0.157 & 0.145  \\
Diagonal-5  & \xmark & \cmark & 0.234 & 0.370 & 0.261 & 0.319 & 0.142  \\
Diagonal-25 & \xmark & \cmark & 0.176 & 0.850 & 0.271 & 0.356 & 0.149  \\
\midrule
BERT  & \cmark & \xmark & 0.142 & 0.056 & 0.075 & 0.158 & 0.149 \\
BLEU1 & \cmark & \xmark & 0.196 & 0.122 & 0.123 & 0.317 & 0.153 \\
BLEU4 & \cmark & \xmark & 0.296 & 0.134 & 0.163 & 0.277 & 0.150 \\
TFIDF & \cmark & \xmark & \textbf{0.428} & 0.173 & 0.229 & 0.367 & 0.152  \\
\midrule
Chronological-Random & \xmark & \cmark & 0.207 & 0.387 & 0.264 & 0.262 & 0.148  \\
Chronological-BERT  & \cmark & \cmark & 0.196 & 0.364 & 0.250 & 0.253 & 0.146  \\
Chronological-BLEU1 & \cmark & \cmark & 0.240 & 0.455 & 0.308 & 0.276 & 0.150  \\
Chronological-BLEU4 & \cmark & \cmark & 0.262 & 0.492 & 0.336 & 0.366 & 0.152  \\
Chronological-TFIDF & \cmark & \cmark & 0.351 & 0.670 & \textbf{0.452} & \textbf{0.407} & \textbf{0.154}  \\
\bottomrule
\end{tabular}
\vspace{-2.5mm}
\caption{\label{font-table} Alignment performance and task accuracy for various methods.}
\vspace{-4mm}
\end{table*}

\subsection{Benchmark NLP Tasks}

We show the benefits of having an alignment by constructing two benchmark NLP tasks from the corpus:
Cloze-form question answering and a simplified form of abstractive summarization.
The alignment $H$ determines the subset of the story document $D$ used for each task.
We choose these tasks to illustrate the benefits of aligned stories and summaries on reading comprehension most effectively.
We expect improved performance given better alignments as the model is not required to search through a long text.
Note that our key contribution is not a new task but rather a different approach that incorporates alignments to improve story understanding.
Additionally, our dataset also serves as a benchmark for unsupervised learning of alignments.



\vspace{1mm}
\noindent\textbf{Cloze-Form question answering.}
We construct this task following~\citet{hermann2015teaching}.
A question $q_{i}$ is constructed by masking an entity in a summary paragraph $s_{i}$. 
Given the subset of story paragraphs from $D$ aligned to $s_{i}$ as determined by $H$, the task is to select the correct answer $a_{i}$ from a set of candidate answers -- up to 10 other entities from the same story.
We determine entities using an entity tagger from AllenNLP~\cite{Gardner2017AllenNLP} and use entity anonymization~\citet{hermann2015teaching} to ensure that the model does not learn information about specific entities.
For a small percentage of $s_{i}$ that do not have entities, we mask a random token during training.
For evaluation, we report accuracy only on questions with masked entities.

Our model is a modified version of the Attentive Reader (AR) from~\citet{hermann2015teaching}, adapted for multi-paragraph texts.
We treat (potentially multiple) story paragraphs aligned to the sentence $s_i$ as a unique document and compute its question-conditioned representation.
This is calculated as an attention-weighted sum of the word vectors, where attention is computed with respect to the question representation.
We then combine individual paragraph representations into a single feature using another attention-weighted sum, again, attended by the question representation.
This essentially amounts to a hierarchical version of AR.
More model details are provided in~Appendix \ref{sec:cloze_model_details}.

\vspace{1mm}
\noindent\textbf{Multi-choice abstractive summarization.}
To see the benefits of having alignments, we propose a simplified form of abstractive summarization as another task.
Given the story document $D$ and corresponding alignments $H$, the goal of this task is to complete summary sentences $s_i$ given first $T=5$ tokens.
In particular, we make this a multiple-choice problem by creating a set of 10 candidate sentences, 1 correct, and 9 incorrect ones obtained from tokens from other summary paragraphs of the same chapter.


Our model is an LSTM decoder that uses a modified version of the Attentive Reader to attend to the aligned portion of the document at each time step.
In particular, we sequentially predict words using an LSTM decoder (similar to other generative summarizers) and rank possible candidate answers by computing the average log-likelihood for each candidate.
The use of multiple-choice answers for summarization removes additional ambiguity introduced by sentence comparison metrics (\emph{e.g.}, BLEU, ROUGE).
Nevertheless, please note that our LSTM decoder is able to generate a summary paragraph when no candidates are provided.
Additional details are in Appendix~\ref{sec:align_model_details}.

\subsection{Learning Alignments}
We also experiment with learning an alignment simultaneously with a task of interest (\emph{e.g.}, Cloze QA).
This has the potential to obtain ever-increasing performance improvements for both the alignment as well as the task.
As the alignment $H$ is a latent non-differentiable variable, we follow a Concave-Convex Procedure~\cite{yu2009learning}.
The optimization procedure alternates between computing optimal $H$ given model parameters $\theta$ using dynamic programming, and learning $\theta$ given $H$ with a gradient-based optimizer.
More details on the procedure are provided in Appendix~\ref{sec:align_learn_details}.

\section{Results}

Table~\ref{font-table} summarizes experimental results on alignment and NLP tasks studied in this work.

\vspace{1mm}
\noindent\textbf{Alignment performance}
is reported as precision, recall, and F1-score, with each ground-truth aligned $(s_{i}, d_{j})$ pair considered as a sample.
We see that even with our simple methods, it is possible to produce reasonably accurate alignments.
Our best result, using TFIDF pairwise scores with a chronological constraint, attains 0.452 F-score, whereas random alignments are 0.186 or below.
Chronological-TFIDF tends to closely follow the ground-truth and has high recall (see Fig.~\ref{fig:alignments}).

Interestingly, while large-scale corpora models like BERT have been successful on many tasks, story-understanding still poses a significant challenge.
Using sentence representations from a pre-trained BERT model performs similar to random scoring and far worse than word matching: F-score of 0.25 for Chronological-BERT, 0.264 for Chronological-Random and 0.452 for Chronological-TFIDF.
We speculate that this is due to the complexity of stories and summaries (\emph{e.g.}, tracking long-range dependencies, high variance in linguistic style), and the generally limited availability of summary-story pairs in the wild.

\vspace{1mm}
\noindent\textbf{Benchmark NLP Tasks.}
We use accuracy as a metric for both Cloze and summarization tasks.
In general, using an alignment helps improve performance, and alignments that take advantage of the chronological structure perform better on both studied tasks.
Without using any alignment, accuracy on the tasks is at 0.149 and 0.148.

We categorize methods based on whether they have access to the similarity scoring function $g$ and whether they use temporal order (diagonal or DP).
Among methods that do not have access to $g$, non-temporal methods (Random-N) achieve an accuracy of 0.255 and 0.144.
In comparison, methods that use temporal order (Diagonal-N) obtain higher accuracy 0.356 and 0.149, respectively.
Among methods with access to $g$, using temporal order does improve performance, but is less pronounced.
For example, Chronological-TFIDF obtains an accuracy of 0.407 and 0.154, while TFIDF achieves 0.367 and 0.152.

This demonstrates that the chronological order present in the corpus is useful for natural language tasks, in particular when no prior information about the text is available.
A reason for the small improvements on summarization is the task difficulty.
On the other hand, we see that Cloze-form QA is made much more accessible by leveraging alignment information.

\vspace{1mm}
\noindent\textbf{Simultaneously learning alignments.}
Jointly optimizing both $H$ and model parameters for the Cloze task produces alignments with an F-score of 0.355.
While this is more accurate than other methods such as Chronological-BERT, it is lower than the Chronological-TFIDF approach.
Learning alignments with weak supervision remains an active area of research~\cite{zhu2015aligning,raffel2017online,luo2017learning}, and is an interesting task that our corpus facilitates.

\vspace{-1mm}
\section{Conclusion}
\vspace{-0.5mm}
We introduced a corpus of stories and loosely-aligned summaries, as well as a set of benchmark tasks for story-based reading comprehension.
We showed that the corpus's structural properties, in the form of temporal ordering, are key to learning effective representations for these tasks.
This is the core value of our corpus, as it makes the challenge of learning on stories much more accessible.
Beyond the tasks we showed here, we believe this corpus can be built upon to expand to other challenges such as question answering and learning alignment, pushing the envelope of story understanding in multiple domains.
We make the corpus available at \url{https://github.com/achaudhury/shmoop-corpus}.

\vspace{1mm}
\noindent \textbf{Acknowledgments.}
We thank Shmoop for creating an amazing learning resource and allowing us to use their summary data for research purposes.
The project was supported by DARPA Explainable AI (XAI) and NSERC Cohesa.
We thank our Upwork annotators for helping us in annotating the alignments between stories and their summaries.

\section*{Appendix}
\appendix
We present additional details on the benchmark Cloze-form QA and summarization tasks (Appendix~\ref{app:tasks}) and their corresponding models in Appendix~\ref{app:models}.
We also discuss the method we adopt for learning alignments simultaneously with the task in Appendix~\ref{sec:align_learn_details} and present some examples of aligned summary and story paragraphs from our corpora in Appendix~\ref{app:examples}.

\section{Task Construction Details}
\label{app:tasks}
\subsection {Cloze-Form Question Answering}

\begin{table}[h]
\centering
\small
\begin{tabular}{p{24em}}
\toprule
\textbf{20,000 Leagues Under the Sea} \\ \\

\textbf{Question:} \\
{[MASK]} obviously wants to find his Giant Narwhal as well. Only Conseil seems uninterested. \\ \\
\textbf{Multiple Choice Answers:} \\
\textit{a)} \textbf{Aronnax} \hspace{4mm} \textit{f)} Giant Narwhal \\
\textit{b)} Nautilus \hspace{5mm} \textit{g)} Higginson \\
\textit{c)} Conseil \hspace{6mm} \textit{h)} Moby Dick \\
\textit{d)} Ned \hspace{11mm} \textit{i)} Vanikoro \\
\textit{e)} Nemo \hspace{9mm} \textit{j)} Kraken \\
\midrule
\textbf{Oedipus the King} \\ \\

\textbf{Question:} \\
{[MASK]} reenters and demands that anyone with information about the former king's murder speak up. He curses the murderer. \\ \\
\textbf{Multiple Choice Answers:} \\
\textit{a)} Creon \hspace{8mm} \textit{f)} Polybus \\
\textit{b)} Jocasta \hspace{6mm} \textit{g)} Apollo \\
\textit{c)} \textbf{Oedipus} \hspace{4mm} \textit{h)} Sphinx \\
\textit{d)} Teiresias \hspace{4mm} \textit{i)} Corinth \\
\textit{e)} Laius \hspace{9mm} \textit{j)} Thebes \\
\bottomrule
\end{tabular}
\vspace{-2mm}
\caption{\label{example_cloze} Examples of Cloze-form questions derived from summary paragraphs. Our task employs 10 multiple choice options.}
\end{table}

We create Cloze-form questions in the following manner.
For each summary paragraph, candidate entities for masking are determined using the pre-trained AllenNLP entity tagger~\cite{Gardner2017AllenNLP}, using the PERSON, ORGANIZATION and LOCATION tags.
Entities that do not occur in the original text are removed from the candidate set.
If multiple entity candidates are found in a single summary paragraph, the entity that appears least frequently in the summary is chosen -- thus encouraging diversity of answers.
If no entities are present in the summary paragraph, a token is chosen at random to be masked during training, but is not included in reporting the final performance.

\subsection{Multi-choice Abstractive Summarization}

\begin{table}[h]
\centering
\small
\framebox[\linewidth]{
\begin{tabular}{p{23em}}
\textbf{A Christmas Carol} \\ \\

\textbf{Question:} \\
Scrooge throws out his famous....  \\ \\
\textbf{Multiple Choice Answers:} \\
\textit{a)} ... come over for Christmas dinner, but Scrooge isn't having any of it. \\
\textit{b)} ... but what about the whole Jesus's birth thing?\\
\textit{c)} \textbf{... catchphrase - Bah! Humbug!.} \\
\textit{d)} ... guys show up asking for any donations for the poor. \\
\textit{e)} ... the cellar bursts open and out of it comes Marley's ghost!. \\
\end{tabular}
}
\vspace{-2.5mm}
\caption{\label{example_summ} Example of Multi-choice Abstractive summarization. Our task employs 10 multiple choice options, but here we show 5 for brevity.}
\vspace{-1mm}
\end{table}

In this task, our goal is to complete the summary bullet point given the first $T=5$ words of the paragraph, the original story $D$ and an alignment $H$.
Summaries that have less than $T=5$ tokens are not considered in this task (both as correct or contrastive answers).
Instead of generating a new sentence, we wish to complete the sentence by choosing one among 10 possible candidate sentences.
The wrong candidates consist of other summary paragraphs from the same chapter (ignoring the first $T$ tokens).

In particular, for a model that consists of a standard RNN decoder (\emph{e.g.}, LSTM), and we teacher force the entire summary paragraph and train the parameters using cross-entropy loss.
At test time, we seed (teacher-force) the model with the first $T$ tokens of the desired summary paragraph.
Then, we compute the average log-likelihood (probabilities) of all remaining tokens in each candidate to rank them.

\section{Model Details}
\label{app:models}
\subsection {Cloze-Form Question Answering}
\label{sec:cloze_model_details}

We first encode the question $q_i$ (summary paragraph $s_i$ with a blank) and story paragraphs $d_j$ with a word embedding and Bi-LSTM to obtain $q_{i}^{enc}$ and $d_{j}^{enc}$ (Eq.~\ref{eq:1},~\ref{eq:2}).
We then use our modified Attentive Reader $f_{AR}$ to generate a conditional document representation $c_{i}$ based on the encoded mask token representation $q_{i[mask]}^{enc}$, and alignment $H$ (Eq.~\ref{eq:3}).
Finally an answer vector $a_{i}^{*}$ is computed by adding $c_{i}$ to $q_{i[mask]}^{enc}$ and applying a linear layer $W_{f}$ (Eq.~\ref{eq:4}).
This answer vector is compared against candidate answers using cosine similarity, followed by selecting the highest scoring answer.
\begin{eqnarray}
q_{i}^{enc} &=& BiLSTM_{q}(W_{emb}(q_{i})) \, , \label{eq:1} \\
d_{j}^{enc} &=& BiLSTM_{d}(W_{emb}(d_{j})) \, , \label{eq:2} \\
c_{i} &=& f_{AR}(q_{i[mask]}^{enc}, H, D^{enc}) \, , \label{eq:3} \\
a_{i}^{*} &=& W_{f}(q_{i[mask]}^{enc} + c_{i}) \, . \label{eq:4}
\end{eqnarray}

Our modified Attentive Reader $f_{AR}$ first constructs a conditional representation $p_{j}$ for each $d_{j}^{enc}$ based on an attention-weighted sum of token representations.
Attention is computed as a function of the encoding of the mask token from the question $q_{i[mask]}^{enc}$ and token representation $d_{jt}^{enc}$ ($q$ and $d_{jt}$ respectively for brevity) in Eq.~\ref{eq:5} and Eq.~\ref{eq:6}.
We then combine the representations into a single vector with an attention-weighted sum once again to produce the final AR output (Eq.~\ref{eq:7} and Eq.~\ref{eq:8}):
\begin{eqnarray}
\alpha_{jt} &=& \frac{\exp(q^{T}W_{\alpha}d_{jt})}{\sum_t \exp(q^{T}W_{\alpha}d_{jt})} \, ,\label{eq:5} \\
p_{j} &=& \sum_{t} \alpha_{jt} d_{jt} \, . \label{eq:6} \\
\beta_j &=& \frac{\exp(q^{T}W_{\beta}p_{j})}{\sum_{j \in h_i} \exp(q^{T}W_{\beta}p_{j})} \, , \label{eq:7} \\
f_{AR}(q,H,D) &=& \sum_{j \in h_i} \beta_{j} p_{j} \, , \label{eq:8}
\end{eqnarray}
where $j \in h_i$ corresponds to the set $\{j : h_{ij} = 1\}$ described by the latent alignment $H$.

Our model parameters $\theta$ constitute a word embedding layer $W_{emb}$, linear layers $W_{\alpha}$, $W_{\beta}$, $W_{f}$, and the bi-directional LSTMs $BiLSTM_{q}$ and $BiLSTM_{d}$.
Dimensionality of embeddings and LSTM hidden states is chosen to be 200.
The model parameters are trained via the max-margin loss (margin $m = 0.1$) using the Adam optimizer with a learning rate of $10^{-4}$.

\vspace{3mm}
\begin{algorithm}[t!]
	\SetAlgoLined
	\caption{CCCP Optimization}
    Initialize $\theta$ and $H$ \;
	\While{not converged} {
        $H^* = \displaystyle{\argmax_H J_{\theta}(H)}$\;
        $\theta = \displaystyle{\argmin_{\theta} (\max_{H} J_{\theta}(H) - J_{\theta}(H^*))}$\;
	}
	\label{algo:cccp}
\end{algorithm}

\subsection{Multi-choice Abstractive Summarization}
\label{sec:align_model_details}
We adapt the Attentive Reader from above to the generation task by computing new context vectors $c_{it}$ for each generated token rather than a single $c_{i}$.
These are used to create an answer prediction vector $a_{it}^{*}$ at each time step.
\begin{eqnarray}
c_{it} &=& f_{AR}(q_{i,t-1}, h_{i}, D^{enc}) \, , \\
a_{it}^{*} &=& W_{f}LSTM_{g}(q_{it}, c_{it}) \, .
\end{eqnarray}

Different from the model used for Cloze-form QA, we use the cross-entropy loss at each time step.
We use a word embedding and LSTM hidden state dimension of 200, and train our model parameters using the Adam optimizer at a learning rate of $10^{-4}$.

\begin{table}[t!]
\centering
\small
\framebox[\linewidth]{
\begin{tabular}{p{23em}}
\large{\textbf{A Christmas Carol}} \\ \\

\textbf{Summary} \\
$\mathbf{s_{1}}$: Scrooge snorts himself awake, and again it's about to be one o'clock. Scrooge is hip to all this now, though, so he doesn't freak out. \\ \\

\textbf{Original Text} \\
$\mathbf{d_{1}}$: AWAKING in the middle of a prodigiously tough snore, and sitting up in bed to get his thoughts together, Scrooge had no occasion to be told that the bell was again upon the stroke of One. He felt that he was restored to consciousness in the right nick of time, for the especial purpose of holding a conference with the second messenger dispatched to him through Jacob Marley's intervention. But finding that he turned uncomfortably cold when he began to wonder which of his curtains this new spectre would draw back, he put them every one aside with his own hands; and lying down again, established a sharp look-out all round the bed. For he wished to challenge the Spirit on the moment of its appearance, and did not wish to be taken by surprise, and made nervous. \\
\end{tabular}
}
\vspace{-2mm}
\caption{\label{example_text2} Aligned summary and stories paragraphs from \textit{A Christmas Carol}}
\vspace{-2mm}
\end{table}

\begin{table}[t!]
\centering
\small
\framebox[\linewidth]{
\begin{tabular}{p{23em}}
\large{\textbf{A Midsummer Night's Dream}} \\ \\

\textbf{Summary} \\
$\mathbf{s_{3}}$: Quince announces that Bottom is the paramour of a sweet voice, and Flute points out that he means "paragon." \\ \\
\textbf{Original Text} \\
$\mathbf{d_{7}}$: QUINCE  Yea, and the best person too, and he is a very paramour for a sweet voice.  \\
$\mathbf{d_{8}}$: FLUTE You must say “paragon.” A “paramour” is (God bless us) a thing of naught. \\ \\
\hline \\
\textbf{Summary} \\
$\mathbf{s_{4}}$: Snug enters the house, announcing that the Duke is coming from the temple with two or three more couples who were just married.  Flute laments that, had they been able to perform, they'd no doubt be rich men, earning them at least sixpence a day (a royal pension). \\ \\
\textbf{Original Text} \\
$\mathbf{d_{9}}$: \textit{Enter Snug the joiner.} \\
$\mathbf{d_{10}}$: SNUG Masters, the Duke is coming from the temple,	 and there is two or three lords and ladies more married. If our sport had gone forward, we had all been made men. \\
$\mathbf{d_{11}}$: FLUTE O, sweet bully Bottom! Thus hath he lost sixpence a day during his life. He could not have ’scaped sixpence a day. An the Duke had not given him sixpence a day for playing Pyramus, I’ll be hanged. He would have deserved it. Sixpence a day in Pyramus, or nothing! \\
\end{tabular}
}
\vspace{-2mm}
\caption{\label{example_text3} Aligned summary and stories paragraphs from \textit{A Midsummer Night's Dream}}
\end{table}

\begin{table}[t!]
\centering
\small
\framebox[\linewidth]{
\begin{tabular}{p{23em}}
\large{\textbf{The Masque of Red Death}} \\ \\

\textbf{Summary} \\
$\mathbf{s_{2}}$: Prince Prospero, the ruler of said kingdom currently being ravaged by the Red Death, is "happy" and "dauntless" and decides he doesn't want to bother with the disease. So he takes a thousand of his knights and maidens and shuts himself up with them in a hidden "castellated abbey" (that would be an abbey made over into a castle, with battlements). \\
$\mathbf{s_{3}}$: The doors of the abbey are welded shut, so no one can get in. But no one can get out, either. \\
$\mathbf{s_{4}}$: Prince Prospero is quite the party animal, and plans to have a good time while the rest of the world dies. \\
$\mathbf{s_{5}}$: The abbey (which Prospero designed himself) is filled to the brim with all the makings of an incredible party: lots of food, jesters, dancers, musicians, and wine. \\ \\
\textbf{Original Text} \\
$\mathbf{d_{2}}$:But the Prince Prospero was happy and dauntless and sagacious. When his dominions were half depopulated, he summoned to his presence a thousand hale and light-hearted friends from among the knights and dames of his court, and with these retired to the deep seclusion of one of his castellated abbeys.  This was an extensive and magnificent structure, the creation of the prince's own eccentric yet august taste.  A strong and lofty wall girdled it in. This wall had gates of iron.  The courtiers, having entered, brought furnaces and massy hammers and welded the bolts.  They resolved to leave means neither of ingress nor egress to the sudden impulses of despair or of frenzy from within.  The abbey was amply provisioned.  With such precautions the courtiers might bid defiance to contagion.  The external world could take care of itself.  In the meantime it was folly to grieve, or to think.  The prince had provided all the appliances of pleasure.  There were buffoons, there were improvisatori, there were ballet-dancers, there were musicians, there was Beauty, there was wine.  All these and security were within.  Without was the "Red Death". \\ 
\end{tabular}
}
\vspace{-2mm}
\caption{\label{example_text4} Aligned summary and stories paragraphs from \textit{The Masque of Red Death}}
\end{table}

\section{Learning Alignment Details}
\label{sec:align_learn_details}

We provide additional details for our approach towards learning the alignment jointly with the task.

Let $J_{\theta}(H)$ be the loss function for a given task, for a model with parameters $\theta$, depending on a latent variable $H$.
To jointly optimize both $\theta$ and $H$ we follow a modified version of the Concave-Convex Optimization Procedure (CCCP) outlined in Algorithm~\ref{algo:cccp}.


Notably, the initial step of solving $H^*$ can be computed efficiently due to the alignment constraints with a dynamic programming algorithm as $J_{\theta}$ can be made to have a linear dependency on $H$.
For the sake of simplicity use contrastive sampling to estimate the max over all possible alignments during training.
During validation, our focus is on determining $H^*$ and we do not actually compute $\max_{H} J_{\theta}(H)$.

\section{Additional Dataset Examples}
\label{app:examples}

We present several examples of summary paragraphs and their corresponding aligned story paragraphs.
We include an example of a perfect one-to-one paragraph alignment from a novel (\textit{A Christmas Carol}) in Table~\ref{example_text2};
a one-to-many alignment common in the presence of story paragraphs consisting of dialogs in a play (\textit{A Midsummer Night's Dream}) in Table~\ref{example_text3};
and a many-to-one alignment in a large paragraph from a short story (\textit{The Masque of the Red Death}) in Table~\ref{example_text4}.

\bibliography{emnlp-ijcnlp-2019}
\bibliographystyle{acl_natbib}

\end{document}